\documentclass[sigconf]{acmart}
\AtBeginDocument{%
  }

\usepackage{graphicx}
\usepackage{multirow}
\usepackage{amsmath}
\usepackage{natbib}
\usepackage{subfig}
\usepackage{xcolor}
\usepackage{listings}

\copyrightyear{2026}
\acmYear{2026}
\setcopyright{cc}
\setcctype{by}
\acmConference[ICMI '26]{INTERNATIONAL CONFERENCE ON MULTIMODAL INTERACTION}{October 05--09, 2026}{Napoli, Italy}
\acmBooktitle{INTERNATIONAL CONFERENCE ON MULTIMODAL INTERACTION (ICMI '26), October 05--09, 2026, Napoli, Italy}
\acmDOI{10.1145/3776574.3831200}
\acmISBN{979-8-4007-2318-6/2026/10}

\begin{document}

\title{Do Visual Features Improve Other-Initiated Repair Detection? A Dyadic Multimodal Approach}


\author{Anh Ngo}
\affiliation{%
  \institution{ALMAnaCH, INRIA Paris}
  \institution{ISIR, Sorbonne University}
  \city{Paris}
  \country{France}
}
\email{anh.ngo-ha@inria.fr}

\author{Nicolas Rollet}
\affiliation{%
  \institution{ALMAnaCH, INRIA Paris}
  \institution{Télécom Paris, SES, Institut Polytechnique de Paris, I3-CNRS}
  \city{Paris}
  \country{France}
}
\email{nicolas.rollet@inria.fr}

\author{Catherine Pelachaud}
\affiliation{%
  \institution{CNRS, ISIR, Sorbonne University}
  \city{Paris}
  \country{France}
}
\email{catherine.pelachaud@upmc.fr}

\author{Chloé Clavel}
\affiliation{%
  \institution{ALMAnaCH, INRIA Paris}
  \institution{Télécom Paris, LTCI, Institut Polytechnique de Paris}
  \city{Paris}
  \country{France}
}
\email{chloe.clavel@inria.fr}








\begin{abstract}
  Other-initiated Self-repair, or in short Other-initiated Repair (OIR), is an essential mechanism in conversational interaction, whereby a recipient signals a problem in speaking, hearing, or understanding, prompting the previous speaker to resolve it. In the case of conversational agents, it is essential to accurately identify these repair initiation strategies to address communication breakdowns efficiently. While conversational analysis studies have shown that OIR initiation is accompanied by both verbal and non-verbal signals such as gaze shifts, facial expressions, body postures, and hand gestures, existing computational approaches rely mainly on text and audio. This paper introduces a novel multimodal model for OIR detection and classification, incorporating a set of visual features drawn from conversation analysis. We evaluate our approach on two corpora with distinct languages and interaction settings. Results demonstrate that visual information consistently improves performance over text and audio baselines, and provide insights into cross-modal feature contributions across two corpora.
\end{abstract}





\maketitle

\section{Introduction}

Conversational repair is a fundamental mechanism humans use to establish and maintain mutual understanding and resolve communication breakdowns. One of the most common repair mechanisms is \textbf{Other-initiated Self-repair, or Other-initiated Repair (OIR)} \cite{Dingemanse2024}, in which a recipient interrupts the ongoing conversational activity to signal a possible trouble and leaves the prior speaker to identify the trouble source and provide the repair themselves \cite{Schegloff1977, Schegloff2000}. For conversational agents, accurately recognizing repair initiation is crucial for sustaining user engagement and preventing interactional failure \cite{Gehle2014, Arkel2020, Moore2024}. As conversational agents are increasingly deployed in real-world settings, from voice assistants to social robots, robust OIR detection becomes an essential for communicative competence \cite{Ashktorab2019}.

\textbf{Existing computational approaches to OIR recognition have focused almost exclusively on linguistic and acoustic information} \cite{Hohn2017, Purver2018, Alloatti2024, ngo2025}. In contrast, \textbf{Conversation Analysis (CA) has consistently shown that repair initiation is inherently multimodal}, involving coordinated gaze shifts, eyebrow movements, head extensions, gesture holds, and body movement freezes that help interlocutors signal and interpret conversational trouble \cite{Kendrick2015a, Kendrick2015b, McDonough2022, Jokipohja2022, Jokipohja2023}. Despite these findings, \textbf{no previous computational study has investigated how visual behaviors contribute to automatic OIR recognition, leaving an important gap between interactional theory and computational modeling}. Moreover, identical repair expressions can signal different repair types, with their interpretation depending on accompanying non-verbal behaviors \cite{Lutzenberger2024, Homke2025}, highlighting the necessity of integrating visual modality for robust and interpretable OIR detection. This gap motivates our investigation into the role of visual modality in OIR recognition through the following research questions:
\begin{itemize}
\item \textbf{RQ1}: Does incorporating visual modality improve OIR detection and classification?
\item \textbf{RQ2}: How do model performance and visual feature contributions vary across two interaction settings (Table~\ref{tab:corpora})?
\item \textbf{RQ3}: Which groups of visual behaviors contribute most to OIR detection and classification?
\end{itemize}

To address these questions, we present the first computational study to investigate visual behaviors for automatic OIR recognition. \textbf{We introduce a Conversation Analysis (CA)-grounded visual feature representation that enables interpretable modeling of repair behaviors.} Our contributions are fourfold: (1) the first multimodal model incorporating visual behaviors for OIR detection and classification; (2) a CA-grounded visual feature set that captures interactionally motivated non-verbal repair cues; (3) a cross-linguistic, cross-context evaluation on French screen-mediated and Dutch face-to-face conversational corpora; and (4) feature-level analyses revealing how visual behaviors contribute to OIR detection and repair-type classification across interaction settings.

\section{Related Work}
\paragraph{\textbf{Computational OIR detection.}}
In an early computational work on OIR, \citet{Hohn2017} proposed a pattern-based chatbot that handles user-initiated repair in text chats between native and non-native German speakers. \citet{Purver2018} extended this with a supervised classifier utilizing turn-level lexical, syntactic, and semantic parallelism features in English. More recently, \citet{Alloatti2024} introduced a hierarchical tag-based annotation system for repair strategies in Italian task-oriented dialogue, and \citet{ngo2025} addressed OIR detection in conversational dialogue using text and audio modalities, reporting the performance improvement by adding handcrafted linguistic and prosodic features. \textbf{Our work addresses this gap by introducing a multimodal model that integrates visual cues for OIR detection and classification. }

\paragraph{\textbf{Visual signals in OIR}}
CA study has analyzed multiple visual patterns that accompany OIR, such as gaze shifting toward the interlocutor occurring within a $\sim$700ms window before verbal OIR \cite{Kendrick2015a}, sustained held gaze is predominant in open class initiators \cite{Lutzenberger2024}, and mutual gaze is significantly associated with other-repairs \cite{Kosmala2025}. Facial expressions include raised eyebrows that may signal hearing trouble \cite{Oloff2018, McDonough2022}, brow furrow marking understanding trouble \cite{Homke2025, Leinonen2023}, and head movements such as head lateral tilts and pokes may co-occur with partial repeats \cite{Seo2010, Leinonen2023}, associated with repair sequences cross-linguistically \cite{Homke2025, Kosmala2025}. At the body level, forward lean marks the suspension of progressivity \cite{Pajo2023} and full-body immobility spans from the trouble source to repair initiation boundary \cite{McDonough2022, Jokipohja2023}. For gesture, repair initiators frequently recycle the trouble source speaker's preceding gesture with recycled hand shape \cite{Oloff2018, Jokipohja2022}. 

\paragraph{\textbf{Multimodal models for behavioral and facial analysis.}}
Recent work has explored MLLMs for fine-grained facial and behavioral analysis. \citet{Bian2024} showed that traditional tools (OpenFace, Py-Feat) provide numerical precision for AUs and gaze while MLLMs enable contextual reasoning, concluding that combining both is preferable to either alone. \citet{Lan2025}
translated precise AU predictions into chain-of-thought descriptions via GPT-4o, improving micro-expression recognition by providing structured reasoning over subtle cues. \citet{Whitehead2025} showed that zero-shot MLLM annotation of posture in collaborative dialogue achieves high reliability for coarse categories but degrades for nuanced behavioral distinctions, which is relevant to OIR, where subtle differences between open-class and restricted initiators depend on temporal cues MLLMs may miss. \textbf{These findings motivate our design choice, that instead of relying on end-to-end MLLM visual reasoning, we extract structured, CA-grounded visual features and integrate them into a task-specific multimodal model, combining precise feature extraction with learned fusion of temporal patterns across repair sequences.}


\section{Data \& OIR Annotation Framework}
\label{sec:data}

\begin{table}[h!]
\centering
\small
\begin{tabular}{lcc}
\toprule
& \textbf{NOXI} & \textbf{CABB-S} \\
\midrule
Language & French & Dutch \\
Setting & Screen-mediated & Face-to-face \\
Task & Role-play dialogue & Object matching \\
Camera view & Frontal & Semi-frontal \\
\midrule
\# OIR instances & 47 & 378 \\
Duration (hours) & 7 & 8 \\
\midrule
\multicolumn{3}{l}{\textit{OIR Type Distribution}} \\
\quad Open Request & 47.8\% & 6.3\% \\
\quad Restricted Request & 37\% & 10.3\% \\
\quad Restricted Offer & 15.2\% & 83.4\% \\
\bottomrule
\end{tabular}
\caption{Overview corpus statistics}
\label{tab:corpora}
\end{table}

We evaluate on two corpora with distinct interaction settings (Table~\ref{tab:corpora}). \textbf{NOXI} \cite{Cafaro2017} is a French screen-mediated dyadic corpus in which an expert and a novice discuss a pre-selected topic. Each participant is recorded from their own video call stream, providing a frontal view of the face and upper body. \textbf{CABB-S} \cite{Rasenberg2022} is a Dutch face-to-face task-oriented corpus, in which participants alternate between director and matcher roles in an object-matching task. Three cameras capture the interaction, including two semi-frontal participant views and one overview view. The task-oriented setting introduces object-directed gaze and task-related gestures that must be distinguished from general interactional signals.

OIR sequences are annotated following \citet{Dingemanse2015}'s CA framework, consisting of a trouble source, repair initiation, and repair solution segments. The CABB-S annotation scheme was developed based on this framework, and the NOXI annotations were subsequently adapted from it. Repair initiations are categorized into three types: \textit{open requests}, \textit{restricted requests}, and \textit{restricted offers}. Annotations are defined at the \textit{segment} level, a short interaction unit aligned with turns or sub-turns \cite{ngo2026}. Both corpora are highly imbalanced, in which OIR instances constitute only a small fraction of all segments, and OIR-type distributions are highly skewed, with different imbalance patterns across the two corpora.

\section{Methodology}
\label{sec:methodology}
\subsection{Overall Architecture}

\begin{figure*}[h]
\begin{center}
\includegraphics[width=0.9\textwidth]{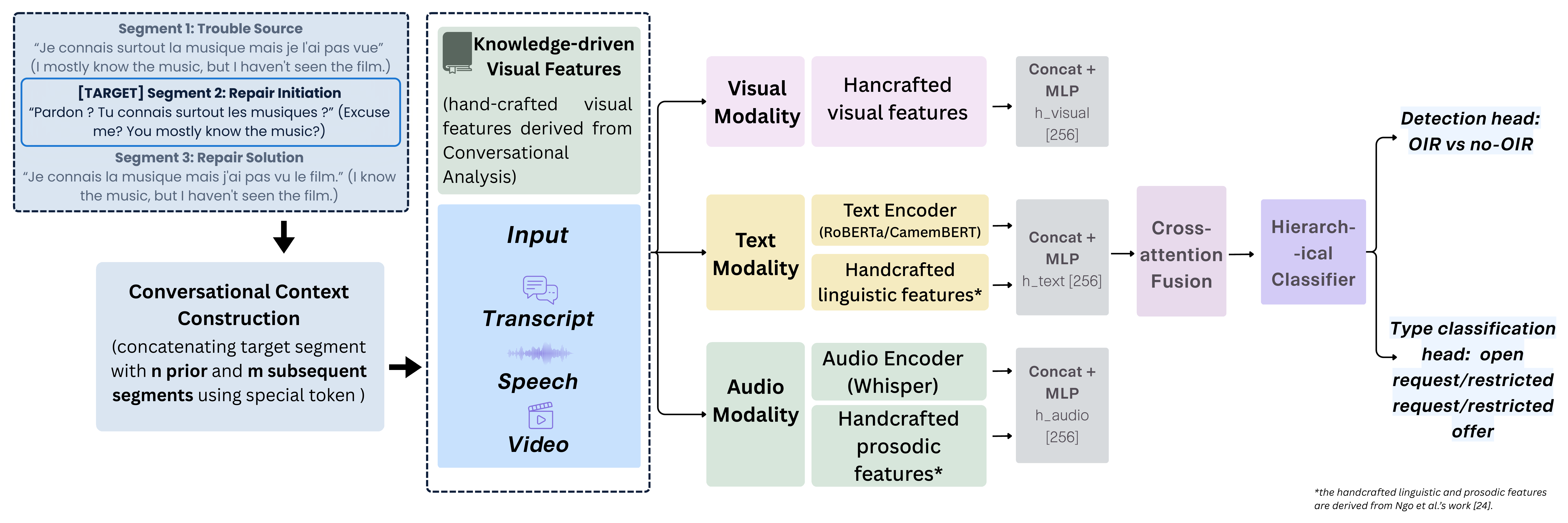}
\caption{Overview of proposed multimodal OIR framework}
\label{fig:overview_multimodal_architecture}
\end{center}
\end{figure*}

Figure~\ref{fig:overview_multimodal_architecture} illustrates the overview of our proposed multimodal model architecture. We adopted and extended the baseline multimodal model from \citet{ngo2025}'s work, by adding the visual branch consisting of handcrafted visual features derived from CA literatures.

\paragraph{Task Formulation.} We frame OIR detection and classification as a segment-level multi-task classification problem. Given a dialogue session segmented into an ordered sequence of segments $S=(s_1,\dots,s_N)$, for each segment $s_i$, the model predicts two layers: (1) a binary detection label  indicating whether the segment is a repair initiation (OIR) or not (Non-OIR), and (2) if it is an OIR, a three-way label classifying the repair initiation type (open request, restricted request, or restricted offer). To reflect this dependency, we utilize a hierarchical classifier in which repair initiation type prediction is conditioned on a positive OIR detection result.

\paragraph{Input Representation.} Inspired by \citet{ngo2025}, who demonstrated the importance of dialogue micro-context, each target segment $s_i$ is concatenated with $n$ preceding and $m$ following segments. The context sizes ($n,m$) are selected specifically for each corpus based on corpus-specific analyses of repair sequence distributions and interaction structure, yielding a context window of $N$ segments. Zero-padding and attention masks handle session boundaries. Each segment is represented by transcript, audio, and video modalities.

\subsection{Text and Audio Branches}
Following \citet{ngo2025}, we utilized pretrained embedding models RobBERT \cite{Delobelle2020} for CABB-S (Dutch) and ModernCamemBERT\cite{Antoun2025} for NOXI (French) as the text encoder, and extracted acoustic representations with the Whisper encoder \cite{Radford2023}. In addition to encoder embeddings, we incorporate similar handcrafted linguistic and prosodic features \cite{ngo2025}. For each modality, handcrafted features are concatenated with encoder output and projected into a shared representation space via a two-layer MLP.

\subsection{Visual Features}

\begin{table}[h]
\centering
\footnotesize
\setlength{\tabcolsep}{3pt}
\renewcommand{\arraystretch}{0.95}
\begin{tabular}{l l p{6cm} l}
\toprule
\textbf{Grp} & \textbf{ID} & \textbf{Feature} & \textbf{C} \\
\midrule
 
\multirow{7}{*}{Gaze}
  & G1 & Gaze direction \cite{Kendrick2015a,Seo2010}. NOXI: toward partner; CABB-S: partner/screen/away (\textit{$9{\times}$bin}). & B \\
  & G2 & Fixation \cite{Lutzenberger2024}: low variance ($\geq$300ms) toward partner. & B \\
  & G3 & Aversion onset \cite{Burton2021}: partner $\rightarrow$ away. & B \\
  & G4 & Eye widening (AU5) \cite{Seo2010}. & N \\
  & G5 & Blink rate (AU45) \cite{Burton2021}: count vs.\ baseline. & N \\
  & G6 & Gaze-to-hands \cite{Mondada2014,Kendrick2015a}: partner $\rightarrow$ hands $\rightarrow$ partner (\textit{$2{\times}$bin}). & B \\
  & G7 & Mutual gaze \cite{Kosmala2025}: dyadic conjunction. & B$^{\sim}$ \\
 
\midrule
 
\multirow{3}{*}{Brows}
  & E1 & Raise (AU1+2) \cite{Oloff2018,Kendrick2015a}: hearing trouble. & B \\
  & E2 & Furrow (AU4) \cite{Homke2025}: understanding trouble. & B \\
  & E3 & Asymmetry $|\text{AU1}_L - \text{AU1}_R|$. & N \\
 
\midrule
 
\multirow{5}{*}{Mouth}
  & M1 & Opening (AU25/26) \cite{McDonough2022} ($\sim$40\% cases). & B$^\dagger$ \\
  & M2 & Lip press (AU24) \cite{Burton2021,Lutzenberger2024}. & B$^\dagger$ \\
  & M3 & Pouch (AU28) \cite{Burton2021,Lutzenberger2024}. & B$^\dagger$ \\
  & M4 & Depressor (AU15) \cite{Kendrick2015a}. & B$^\dagger$ \\
  & M5 & Smile (AU12) \cite{McDonough2022,Kosmala2025}. & B$^\dagger$ \\
 
\midrule
 
\multirow{6}{*}{Head}
  & H1 & Poke \cite{Seo2010,Kendrick2015a}: $\Delta$\texttt{pose\_Tz}. & B \\
  & H2 & Tilt \cite{Seo2010,Pajo2023}: $|\texttt{pose\_Rz}|$. & B \\
  & H3 & Turn \cite{Burton2021,Pajo2023}: $|\texttt{pose\_Ry}|$. & B \\
  & H4 & Nod \cite{Homke2025,Kosmala2025}: oscillation (Rx). & B \\
  & H5 & Shake \cite{Homke2025,Kosmala2025}: oscillation (Ry). & B \\
  & H6 & Velocity: $\ell_2$ norm of pose derivatives. & B \\
 
\midrule
 
\multirow{7}{*}{\shortstack[l]{Body\,\&\\Holds}}
  & B1  & Lean \cite{Kendrick2015a,Pajo2023}: shoulder/hip displacement. & B$^{*}$ \\
  & B2  & Torque \cite{Jokipohja2023}: shoulder--hip angle. & C \\
  & B3a & Hold onset \cite{McDonough2022,Oloff2018} (\textit{one-hot}). & B \\
  & B3b & Hold duration \cite{Leinonen2023} (\textit{$z$-norm}). & B \\
  & B3c & Hold release (\textit{bin}). & B \\
  & B4a & \texttt{hold\_entry} (\textit{bin}). & B \\
  & B4b & \texttt{hold\_exit} (\textit{bin}). & B \\
 
\midrule
 
\multirow{6}{*}{Gesture}
  & Ge1 & Onset latency \cite{Jokipohja2022} (\textit{$z$ + bin $<$800ms}). & B \\
  & Ge2 & Hand shape \cite{Oloff2018,Lutzenberger2024} (\textit{$12{\times}$bin}). & B \\
  & Ge3 & Gesture hold \cite{Oloff2018,Jokipohja2023}. & B \\
  & Ge4 & Gaze--gesture coordination \cite{Jokipohja2022}. & B \\
  & Ge5 & Recycling (DTW) \cite{Jokipohja2022} (\textit{cont + int + bin}). & B \\
  & Ge6 & Palm orientation (cosine $>$0.5). & B \\
 
\bottomrule
\end{tabular}
\caption{Visual features (speaker \& recipient). \textbf{C:} B=both; N=NOXI; C=CABB-S;
$^{*}$ limited NOXI; $^\dagger$ limited CABB-S; $^{\sim}$ approx.\ NOXI.}
\label{tab:visual_features_summarize}
\end{table}

We extract visual features from both participants using OpenFace 2.0 \cite{Baltrusaitis2018} (facial Action Units (AUs) \cite{Ekman1978}, head pose, gaze) and MediaPipe Holistic \cite{Lugaresi2019} (upper-body and hand landmarks). For each target input segment, we derive three feature sets: (1) \textbf{current speaker features}, capturing the speaking participant’s visual behaviors; (2) \textbf{recipient features}, capturing the non-speaking participant’s behaviors over the same time window, encoding possible visible sign of having trouble before taking the floor \cite{Kendrick2015a, Burton2021, Kosmala2025}; and (3) \textbf{interactive features}, encoding dyadic patterns such as mutual gaze. This design allows the transformer-based model to implicitly capture cross-speaker interaction dynamics.

Table~\ref{tab:visual_features_summarize} summarizes visual features and their corpus availability, grounded in CA studies of non-verbal OIR signals, covering gaze \cite{Kendrick2015a, Kendrick2015b, Burton2021}, facial displays \cite{McDonough2022, Homke2025}, head movement \cite{Seo2010, Leinonen2023}, body movement holds \cite{Jokipohja2023, Oloff2018}, body posture \cite{Kendrick2015b, Pajo2023}, and hand gesture \cite{Oloff2018, Jokipohja2022}. As these signals are temporally dynamic, the same feature might express different interactional meaning depending on "when" within a turn it occurs, e.g., the eyebrows and gaze shifts in \citet{Homke2025}'s study. To capture this without requiring frame-level sequence and to smooth out fine-grained noise, we divided the given speaker segment into three equal windows (early/mid/late). Frame-level values are aggregated per window using three strategies depending on the feature type: (1) mean intensity for continuously-valued signals (e.g., AU activations, head pose); (2) proportion of frames above threshold for sustained states (e.g., gaze direction, movement holds); and 3) a binary presence flag indicating whether a transient event occurred (e.g., head nods, head hakes). All continuous features are z-normalized against each speaker's per-session baseline (mean $\mu_s$, standard deviation $\sigma_s$ computed over all frames in the session), so that the binarization threshold $\theta_s = \mu_s + 0.5\sigma_s$ reflects deviation from that individual's own typical behavior rather than a corpus-wide norm. Detailed feature specifications are in the Appendix.

\section{Experiments}
\label{sec:experiment}

\subsection{Experimental Configurations}  
To address our research questions, we evaluate four experiment configurations on both NOXI and CABB-S


\paragraph{Zero-shot MLLM baseline.} We evaluate Qwen2.5-Omni-7B \cite{Xu2025} \footnote{\url{https://huggingface.co/Qwen/Qwen2.5-Omni-7B}} in a zero-shot setting as a reference for end-to-end multimodal reasoning without task-specific training. The model receives dialogue transcripts, audio, and video frames of both participants for the target segment and its surrounding context, together with a structured prompt. This baseline assesses whether a general-purpose MLLM can capture complex OIR interactional phenomena without task-specific fine-tuning.

\paragraph{Text + audio baseline.}$\text{Baseline}_{\text{text+audio}}$\citep{ngo2025} is a multimodal model based on text and audio, serving as the reference for assessing the contribution of visual modality (RQ1). While the original study evaluates it on a class-balanced subset of CABB-S, we instead use the original unbalanced distributions for both corpora to better reflect the nature of the data.

\paragraph{Visual-only model.} $UniVis$ is a unimodal model using only the handcrafted visual features (Table~\ref{tab:visual_features_summarize}), evaluating the standalone contribution of visual cues.

\paragraph{Multimodal model.} $TriModal$ is our proposed multimodal model combining text, audio, and visual modalities, including the complete dyadic visual feature set. 

\subsection{Implementation Details}
\label{sec:implementation_details}

\paragraph{Training Details.} We fine-tune our models using PyTorch Lightning with the AdamW optimizer (weight decay $0.01$). We tuned the learning rate over multiple values and selected $2 \times 10^{-5}$ for experiment configurations with pretrained language and audio encoders, and $1 \times 10^{-3}$ for the visual-only unimodal model. We applied a ReduceLROnPlateau scheduler to monitor the validation macro-F1 score and reduce the learning rate after 3 epochs without improvement. Training runs for up to 30 epochs with early stopping of $patience = 5$. To address the high class imbalance in the detection task across both corpora, we use BCEWithLogitsLoss with a positive class weight defined as  $\mathit{pos\_weight} = N_{\text{neg}} / N_{\text{pos}}$, computed per training fold. For the OIR type classification task, we apply class weights derived from training-fold frequencies in CrossEntropyLoss. The source code is publicly available \footnote{\url{https://github.com/haanh764/multimodal_repair_recognition}}

\paragraph{Cross-Validation.} All conversation sessions are first shuffled with a fixed random seed. We then perform a conversation session-level k-fold cross-validation with $k = 5$ to ensure that no session appears in multiple splits and eventually prevent data leakage. In each fold, 20\% of the sessions are held out as the test set. We further reserve 20\% from the remaining training data as a validation set for early stopping and checkpoint selection. Feature normalisation statistics (mean and standard deviation) are computed exclusively on the training sessions of each fold and applied to the validation and test sets, preventing any leakage of distributional information across splits. Final performance is reported as the mean and standard deviation of the test-set metrics across all five folds.

\paragraph{Evaluation Metrics.} For both OIR detection and type classification, we report  precision, recall, and macro F1-score, in which the primary metric is macro-F1, which assigns weight equally to each class regardless of its frequency, making it suitable for the severe class imbalance in both corpora.

\section{Results \& Discussion}
\label{sec:result}

\begin{table*}[ht]
\centering
\small
\setlength{\tabcolsep}{3pt} 
\begin{tabular}{l ccc ccc ccc ccc}
\toprule
& \multicolumn{6}{c}{\textbf{Detection}}
& \multicolumn{6}{c}{\textbf{Classification}} \\
\cmidrule(lr){2-7} \cmidrule(lr){8-13}
& \multicolumn{3}{c}{NOXI} & \multicolumn{3}{c}{CABB-S}
& \multicolumn{3}{c}{NOXI} & \multicolumn{3}{c}{CABB-S} \\
\cmidrule(lr){2-4} \cmidrule(lr){5-7}
\cmidrule(lr){8-10} \cmidrule(lr){11-13}
Model
& P & R & F1 & P & R & F1 & P & R & F1 & P & R & F1 \\
\midrule
$Qwen2.5-7B-Omni$
& 50.1 & 55.2 & 35.4
& 41.1 & 54.0 & 54.1
& 33.3 & 25.8 & 29.1
& 31.1 & 37.6 & 32.5 \\

$\text{Baseline}_\text{text+audio}$
& 62.9±1.7 & \textbf{68.4±4.1} & 67.0±2.3
& 73.4±6.2 & 66.7±4.9 & 69.2±3.6
& 39.0±1.5 & 43.4±1.2 & 55.3±2.0
& 59.1±12.1 & 57.2±6.5 & 65.3±6.9 \\

$UniVis$
& 57±0.4 & 60±10.0 & 63.9±0.7
& 61.9±11.3 & 60±4.6 & 69.5±2.2
& \textbf{49.5±0.4} & \textbf{61±3.3} & 49.3±0.9
& 48.5±7.5 & 44.7±0.6 & 57.5±5.5 \\

$TriModal$
& \textbf{73.3±4.7} & 53.5±1.5 & \textbf{71.1±8.2}
& \textbf{80.1±6.4} & \textbf{79±3.0} & \textbf{79.6±1.9}
& 43.6±8.5 & 52±10.7 & \textbf{68.5±13.2}
& \textbf{62.5±0.4} & \textbf{59.5±0.4} & \textbf{67.7±3.4} \\
\bottomrule
\end{tabular}
\caption{Main results for OIR detection (binary: OIR vs.\ Non-OIR) and fine-grained classification (P: Precision, R: Recall, F1: macro F1 score) on NOXI and CABB-S.}
\label{tab:results}
\end{table*}

Table~\ref{tab:results} presents detection and classification results across both corpora. We discuss findings organized around our three research questions.

\subsection{Visual Modality Contribution Across Corpora (RQ1 \& RQ2)}  

To answer research questions \textbf{RQ1} and \textbf{RQ2}, we evaluate whether incorporating visual features improves OIR detection and classification (RQ1) and how these gains vary across the two corpora (RQ2). We first compare the proposed \textit{TriModal} model against the text-audio \textit{Baseline} to quantify the benefit of incorporating visual features, and then compare \textit{UniVis} to assess the contribution of visual features alone.

\paragraph{\textbf{OIR Detection.}} Firstly, compared with the zero-shot MLLM, all task-specific models achieve substantially better performance, indicating that general multimodal reasoning is insufficient for OIR detection. Qwen2.5-Omni-7B struggles on both corpora, with 35.4\% and 54.1\% F1-scores for NOXI and CABB-S, respectively. To evaluate the contribution of CA-grounded visual features, we compare the proposed \textit{TriModal} model against the \textit{Baseline} model trained on text and audio input. It could be seen that incorporating handcrafted visual features consistently improves F1-score by 4.1 percentage points (pp) on NOXI and 12.9 pp on CABB-S. The larger improvement on CABB-S might be explained by its task-oriented setting, where interaction patterns are more regular and visually grounded (e.g., repeated gestures during object description), making them easier for the model to exploit. On CABB-S, visual features improve both precision (+6.7 pp) and recall (+12.3 pp), indicating fewer false positives and more correctly detected OIR instances. Meanwhile, on NOXI, they mainly improve precision (+10.4 pp) at the expense of recall (-14.9 pp), suggesting a more conservative detector. This effect is likely amplified by the highly imbalanced and smaller number of OIR instances in NOXI, making recall more sensitive. The \textit{UniVis} model further confirms that the handcrafted visual features themselves carry meaningful information, achieving performance comparable to the baseline (63.9\% on NOXI and 69.5\% on CABB-S) with relatively balanced precision-recall trade-offs. However, the remarkably lower precision compared ($-16.3$ pp vs.\ baseline, $-18.2$ pp vs.\ $TriModal$) indicates that visual features alone lack sufficient specificity and benefit more from fusion with text and audio. 

\paragraph{\textbf{OIR Type Classification.}} Similar to detection task performance, the zero-shot MLLM performs near chance level on both corpora, confirming that fine-grained repair initiation type distinctions cannot be resolved by zero-shot multimodal reasoning without task-specific training. Compared to the baseline, the gain patterns differ notably from the detection task, in which $TriModal$ improves F1-score by 13.2 pp on NOXI and only 2.4 pp on CABB-S. The larger gain on NOXI suggests that visual cues provide discriminative information across OIR types that text and audio alone cannot capture in screen-mediated dialogues. In contrast, on CABB-S, the task-oriented structure produces lexically explicit repair types, which are already well captured by text and audio, leaving less room for improvement. The $UniVis$ model performs worse than the baseline on both NOXI and CABB-S, confirming that visual features alone are insufficient for type disambiguation. Moreover, on NOXI, the lower precision suggests that visual cues alone may overgeneralize across classes, likely due to the small number of examples per class, making class boundaries harder to learn. This highlights the importance of multimodal fusion.

\paragraph{\textbf{Result stability.}} Additionally, we computed the standard deviation across folds for all metrics. Across all configurations, CABB-S results are consistently more stable than NOXI. For $TriModal$, detection variance is substantially lower on CABB-S compared to NOXI, and the patterns are more prominent for classification (CABB-S: $67.7 \pm 3.4$ vs.\ NOXI: $68.5 \pm 13.2$), which could be explained by two factors. First, CABB-S contains more OIR instances with an inherently trial-based structure, which resulted in a more uniform distribution of OIR instances across sessions. Meanwhile, NOXI’s open-ended conversation might lead to uneven distributions, resulting in some sessions containing more repair instances than others. Second, on NOXI, the extremely small number of restricted offers makes per-class F1 dramatically sensitive to whether these instances appear in a given fold, resulting in high variance. The result implies that the variability reflects properties of the data rather than a model limitation, and the wide confidence intervals on NOXI classification results should be interpreted accordingly.

\subsection{Visual Feature Importance Analysis (RQ3)}  
To analyze the feature contribution towards the model decision, we examined the following analysis on our proposed $TriModal$ model, including: modality ablation, gradient saliency, SHAP feature importance, and cross-modal co-activation.

\subsubsection{\textbf{Overall modality importance.}}

\begin{figure}[h]
\begin{center}
\includegraphics[width=\columnwidth]{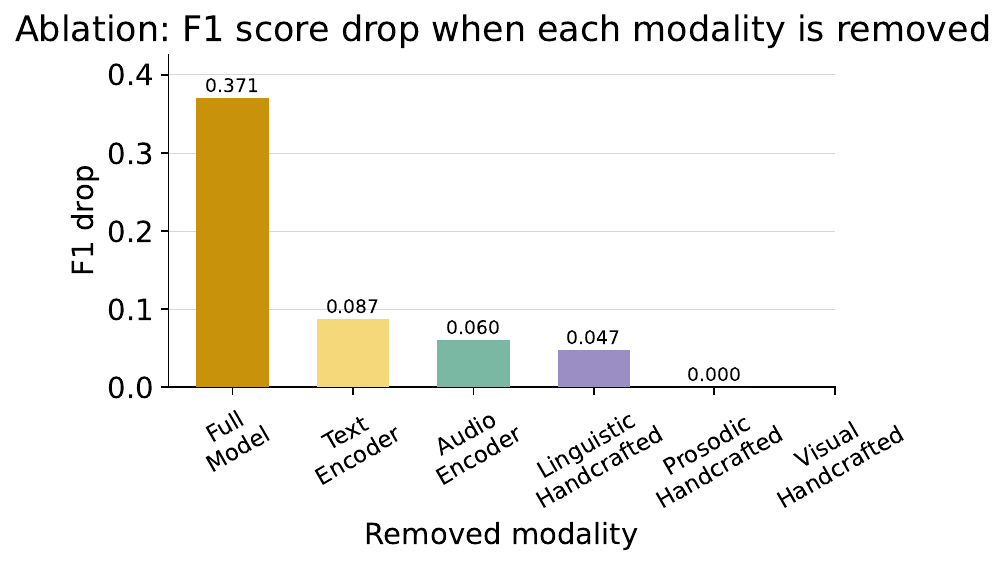}
\caption{NOXI's ablation modality analysis: F1 score drop when each feature group is removed}
\label{fig:noxi_ablation_modality}
\end{center}
\end{figure}

\begin{figure}[h]
\begin{center}
\includegraphics[width=\columnwidth]{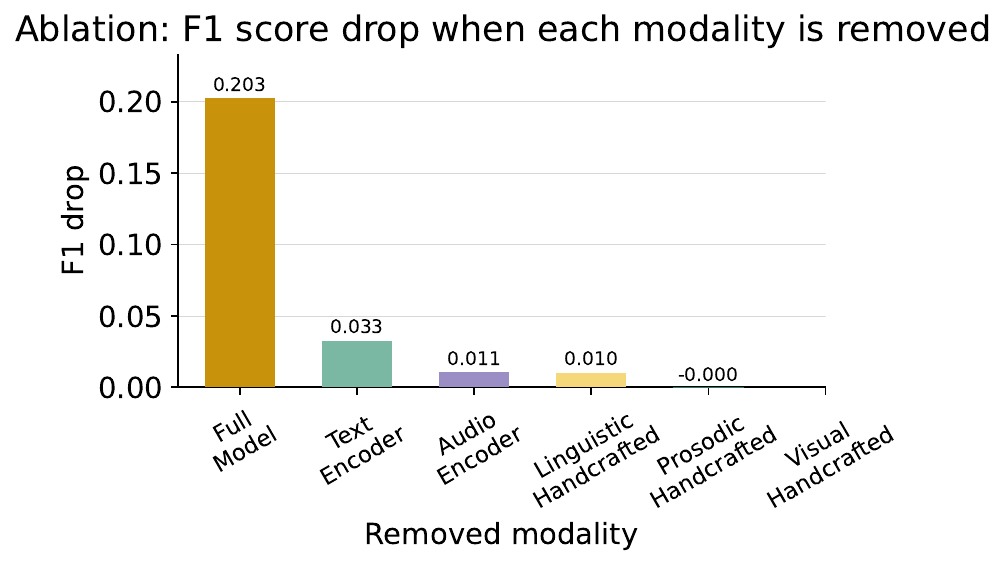}
\caption{CABB-S's ablation modality analysis: F1 score drop when each modality is removed}
\label{fig:cabbs_ablation_modality}
\end{center}
\end{figure}

Figure~\ref{fig:noxi_ablation_modality} and Figure~\ref{fig:cabbs_ablation_modality} illustrate the F1 score drop when each feature group is removed from the multimodal model on NOXI and CABB-S, reflecting not only the impact of modality but also the representation type (learned encoder embeddings vs.\ handcrafted features). In NOXI, removing the text encoder leads to the largest drop (-37.1 pp), confirming that lexical content is the primary signal for OIR detection. Handcrafted linguistic features contribute substantially (-8.7 pp), followed by prosodic (-6 pp) and visual features (-4.7 pp). In CABB-S, the text encoder remains dominant (-20.3 pp), but the contribution of handcrafted features differs, in which prosodic features (-3.3 pp) contribute more than linguistic features (-1.0 pp), with visual features in between (-1.1 pp). This difference reflects how OIR is expressed in each corpus, resulting from the nature of each corpus. In NOXI’s natural conversation, lexical cues such as specific repair tokens are the main signals, while in CABB-S’s task-oriented dialogue, prosodic cues (e.g., pitch variation, intensity) carry more discriminative information, likely because the structured object-matching context makes the speaker’s acoustic patterns more salient. Notably, the raw audio encoder contributes negligibly in both corpora, suggesting that end-to-end acoustic representations add little beyond what prosodic features already capture at this scale.

\begin{figure}[h]
\begin{center}
\includegraphics[width=0.95\columnwidth]{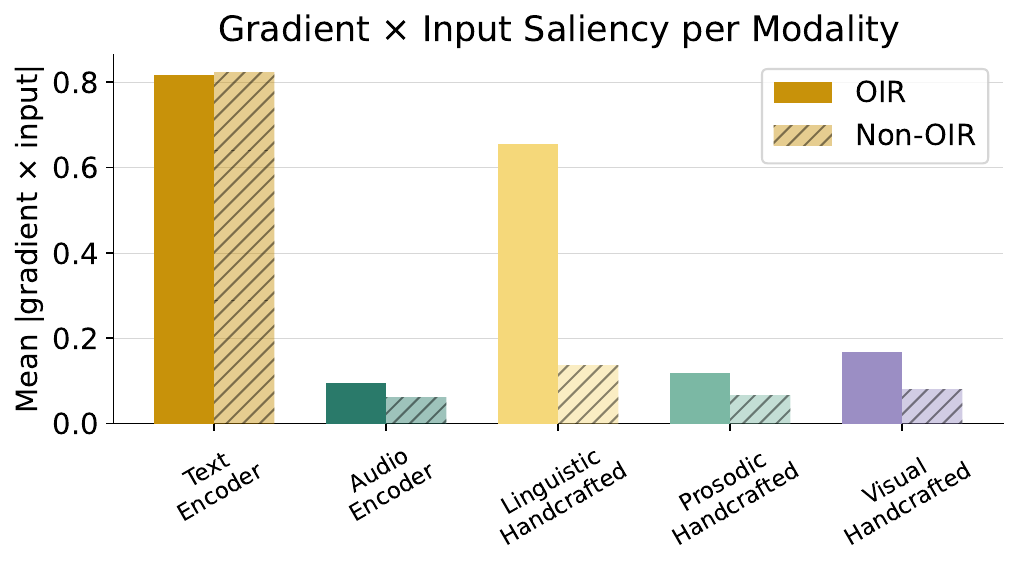}
\caption{NOXI Gradient Saliency}
\label{fig:noxi_grad_saliency}
\end{center}
\end{figure}

\begin{figure}[h]
\begin{center}
\includegraphics[width=0.95\columnwidth]{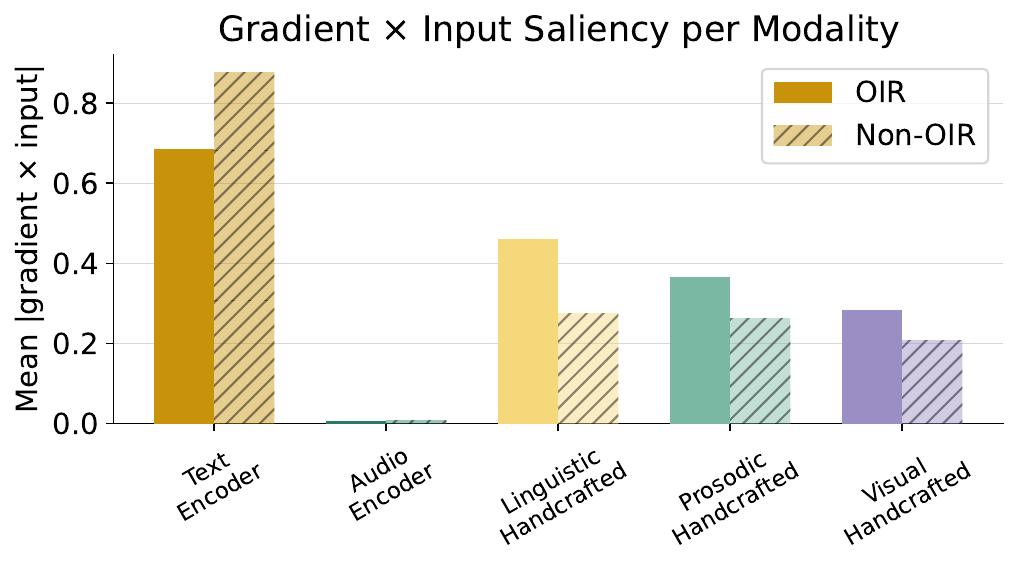}
\caption{CABB-S Gradient Saliency}
\label{fig:cabbs_grad_saliency}
\end{center}
\end{figure}

To further understand which feature group the model relies on when making OIR detection, Figure~\ref{fig:noxi_grad_saliency} (NOXI) and Figure~\ref{fig:cabbs_grad_saliency} (CABB-S) introduce the gradient saliency analysis, which measures how much each input feature influences the model’s output by computing the gradient of the predicted class score with respect to each input \cite{Shrikumar2017}. A high gradient $\times$ input value means the feature is both active and strongly drives the model's decision, and a large gap between OIR and non-OIR segments indicates the feature is more discriminative rather than active equally for all input.  It could be seen in Figure~\ref{fig:noxi_grad_saliency} that for NOXI, linguistic features are the most discriminative for OIR, in which the gap between OIR vs non-OIR is largest for linguistic (0.65 vs.\ 0.13), followed by visual (0.17 vs.\ 0.08) and prosodic (0.12 vs.\ 0.06). Text embeddings exhibit high saliency for both classes, indicating that while the text encoder is highly involved, it is not selectively informative for OIR in distinguishing between OIR and non-OIR compared to the handcrafted features. On CABB-S, the picture is considerably different, in which text embeddings' saliency is notably higher for non-OIR segments than OIR segments, suggesting that the model relies less on textual content during repair initiation rather than during regular conversational turns, which could be explained by the rich object descriptions in regular turns compared to shorter and lexically sparse repair initiation tokens. Linguistic features remain discriminative (OIR 0.46 vs.\ non-OIR 0.28), but prosodic features now show a comparably stronger gap (0.37 vs.\ 0.26), aligning with the reversed handcrafted feature trend observed in the ablation results. Visual features on CABB-S present a higher overall activation level, suggesting that visual behaviours are more broadly active across all segments, not only repair initiations due to the task-driven gestural activity of the object-matching context in CABB-S.

\subsubsection{\textbf{Which visual features matter most.}}

\begin{figure}[h]
\begin{center}
\includegraphics[width=\columnwidth]{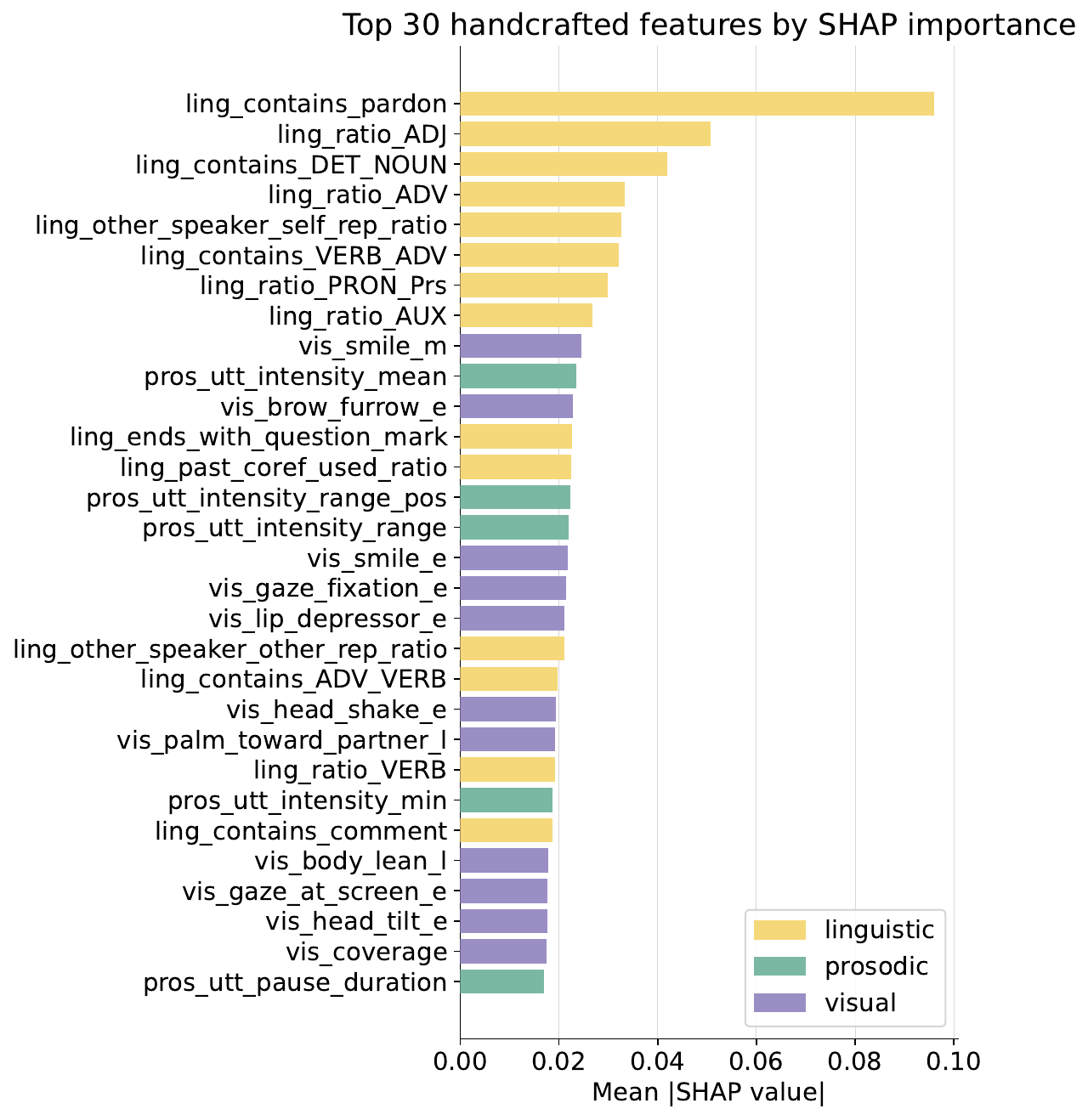}
\caption{NOXI's top 30 most important handcrafted features by SHAP values}
\label{fig:noxi_30_shap}
\end{center}
\end{figure}

Figure~\ref{fig:noxi_30_shap} illustrates the 30 most important handcrafted features for the OIR detection using SHAP values. For NOXI, visual features begin appearing at rank 9 with \textbf{vis\_smile\_m} (current speaker smiles in the middle of their segment) and followed by other facial expressions, such as \textbf{vis\_brow\_furrow\_e} (frowning eyebrow early in the beginning), \textbf{vis\_gaze\_fixation\_e} (current speaker looks at a point steadily in the beginning of his talk), \textbf{vis\_lip\_depressor\_e}, \textbf{vis\_head\_shake\_e} (current speaker begins the segment with head shake), \textbf{vis\_palm\_to-
-ward\_partner\_l} (current speaker’s palm oriented toward partner at the end of their segment), and \textbf{vis\_body\_lean\_l} (current speaker leaning forward) among the most important. It could be seen that early-phase features dominate among these visual patterns, consistent with the CA literature that non-verbal OIR signals onset before the verbal repair initiation \citep{Kendrick2015a, Burton2021}. Moreover, the visual features span multiple groups, including facial displays (brow furrow, smile, lip depressor), gaze (gaze fixation, gaze at screen), head movement (head shake, head tilt), and body posture (body lean, palm orientation), which confirmed that the model learns the whole-body phenomenon, aligned with the CA studies. 

\begin{figure}[h]
\begin{center}
\includegraphics[width=\columnwidth]{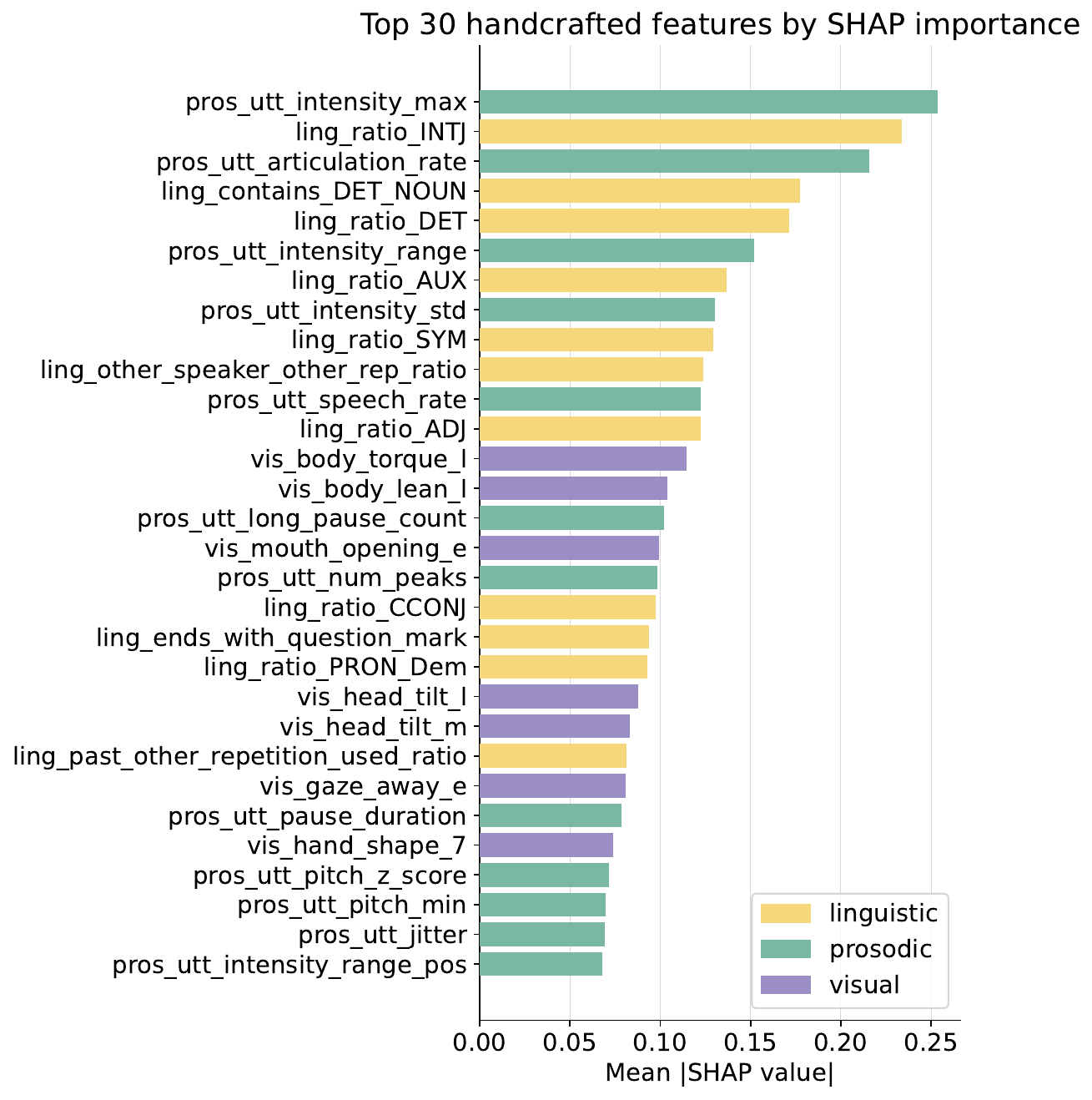}
\caption{CABB-S's top 30 most important handcrafted features by SHAP values}
\label{fig:cabbs_30_shap}
\end{center}
\end{figure}

On CABB-S (Figure~\ref{fig:cabbs_30_shap}), the dominant visual features shift to mainly \textbf{body} and \textbf{head movement}, such as \textbf{body torque} and \textbf{leaning}, and \textbf{head tilting}, with late-phase activations. This shifting could be explained due to the fact that in screen-mediated NOXI, facial signals are primarily visible, so the model can rely on subtle facial muscle expression, while in CABB-S’s semi-frontal camera views can capture well the whole-body posture and head orientation. The late-phase dominance suggests that body-level signals accumulate toward the end of the repair initiation rather than occurring at the onset, which is new temporal patterns not observed in literature.

\subsubsection{\textbf{Cross-modal co-activation analysis}}

\begin{figure}[h]
\begin{center}
\includegraphics[width=\columnwidth]{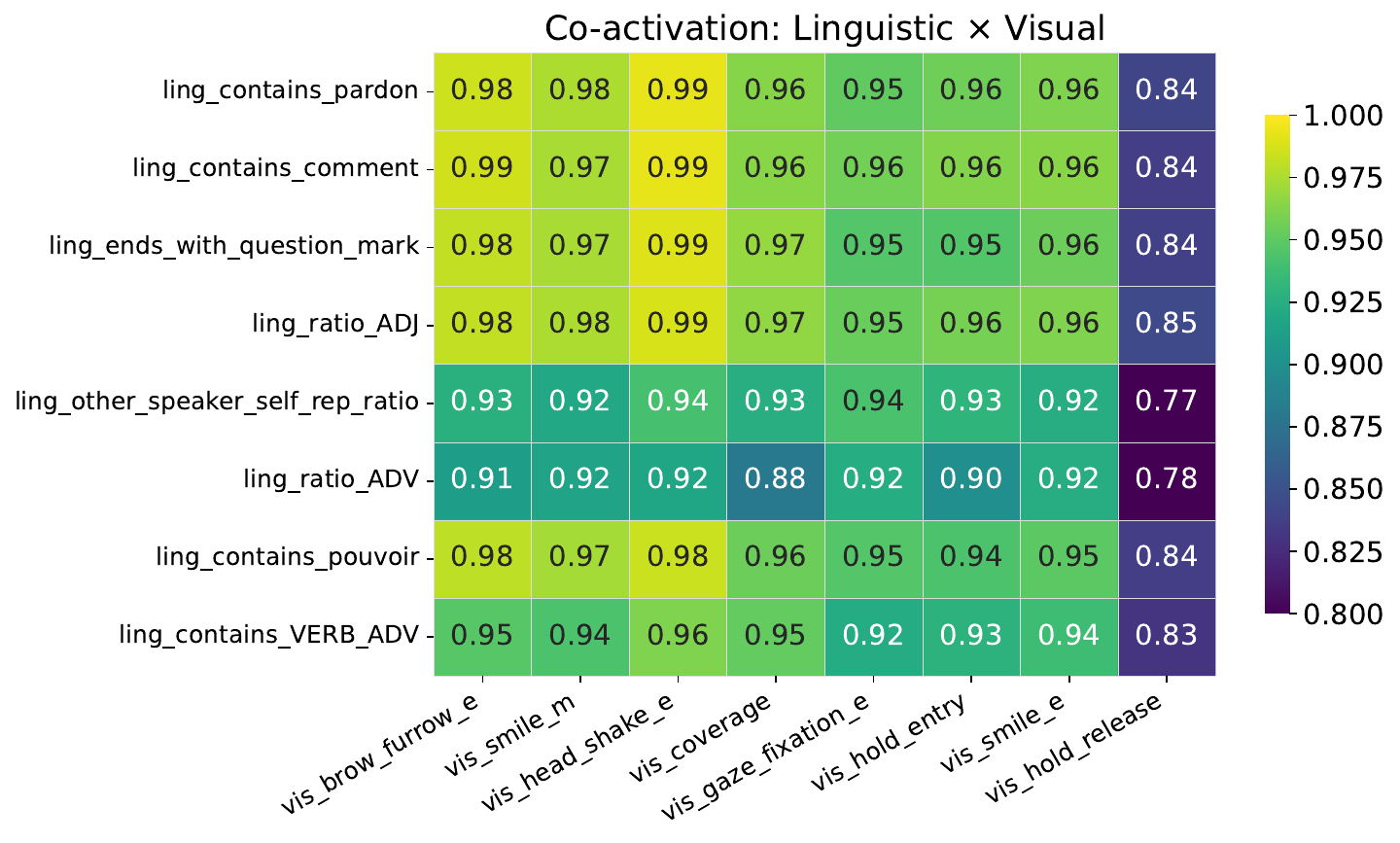}
\caption{NOXI's Linguistic vs Visual co-activation heatmap}
\label{fig:noxi_lin_vis_syn}
\end{center}
\end{figure}

\begin{figure}[h]
\begin{center}
\includegraphics[width=\columnwidth]{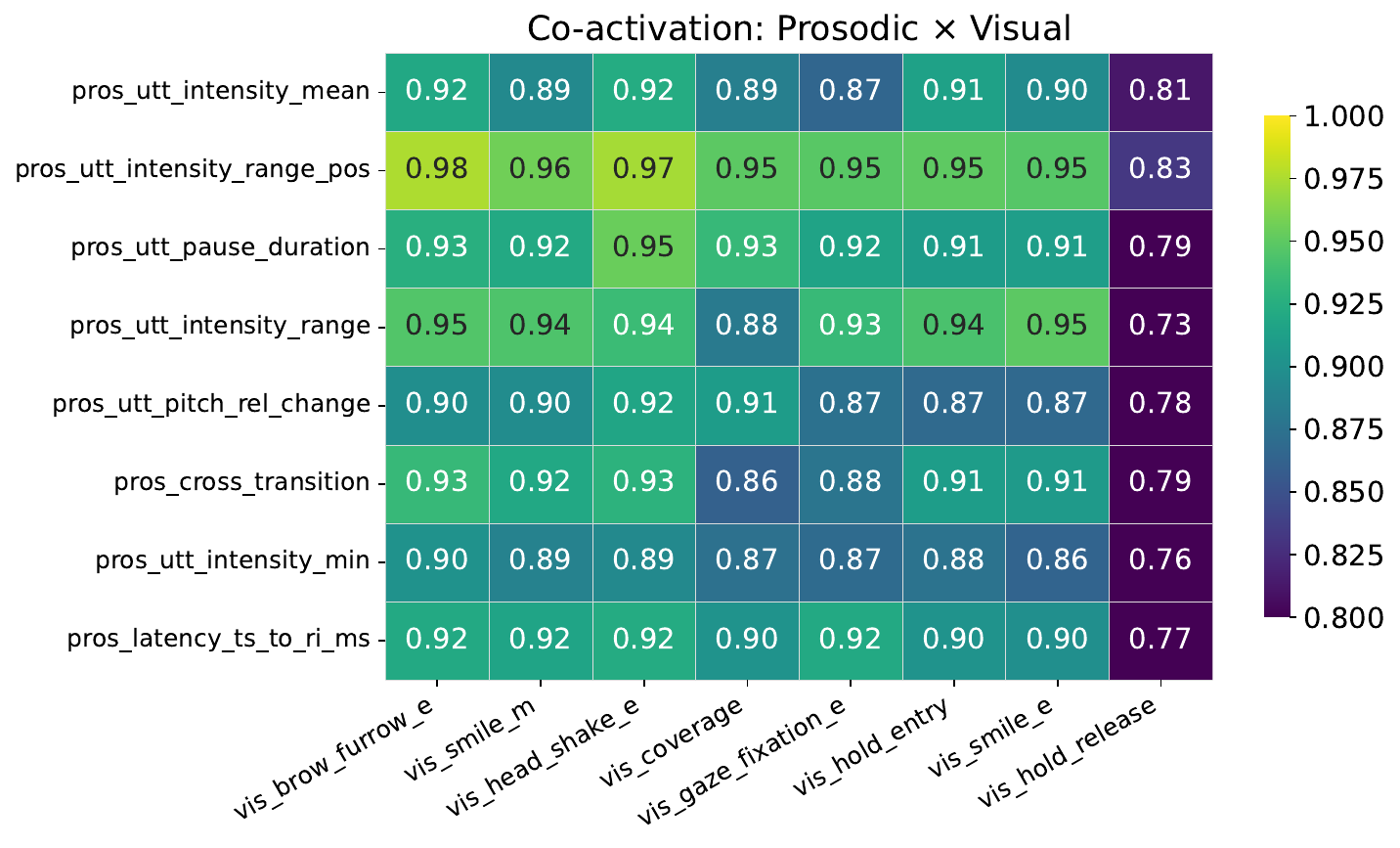}
\caption{NOXI's Prosodic vs Visual co-activation heatmap}
\label{fig:noxi_pros_vis_syn}
\end{center}
\end{figure}

To examine how visual features interact with linguistic and prosodic cues, we compute cross-modal co-activation scores for feature pairs across modalities and visualize co-activation of the top 30 SHAP-ranked features as heatmaps. For each pair of features $(a,b)$ from two different modalities, we define the co-activation score as the Pearson correlation between their per-sample gradient saliency vectors:
\begin{equation}
S(a,b)=\max\!\left(0,\; r(\mathbf{g}_a,\mathbf{g}_b)\right)
\end{equation}
where $\mathbf{g}_a=\left(\left\|\frac{\partial \hat{y}}{\partial \mathbf{h}_a}\odot\mathbf{h}_a\right\|_2\right)_{n=1}^{N}$ is the vector of per-sample gradient$\times$input magnitudes for feature $a$ across the $N$ instances, and $r(\cdot,\cdot)$ denotes the Pearson correlation coefficient. A higher score indicates that the model assigns high importance to both features on the same instances, whereas lower scores indicate that the features are activated more independently. Negative correlations are truncated to zero because our goal is to identify jointly activated cue combinations.

For NOXI, linguistic-visual co-activation (Figure~\ref{fig:noxi_lin_vis_syn}) is generally high (0.84-0.99), indicating that important linguistic and visual cues tend to be activated together. Prosodic-visual co-activation (Figure~\ref{fig:noxi_pros_vis_syn}) shows a similar pattern, with intensity features exhibiting the strongest co-activation with eyebrow furrow activating in the early phase of the segment (\textbf{vis\_brow\_furrow\_e}, $\approx$0.98). In contrast, \textbf{vis\_hold\_release} consistently shows lower co-activation (0.77-0.85), suggesting that the model relies on this visual cue more independently from linguistic and prosodic information. This aligns with CA observations that hold release marks the resumption of movement after a freezing hold during a repair initiation \citep{Jokipohja2023}, which may not be accompanied by either verbal or acoustic patterns.

\begin{figure}[h]
\begin{center}
\includegraphics[width=\columnwidth]{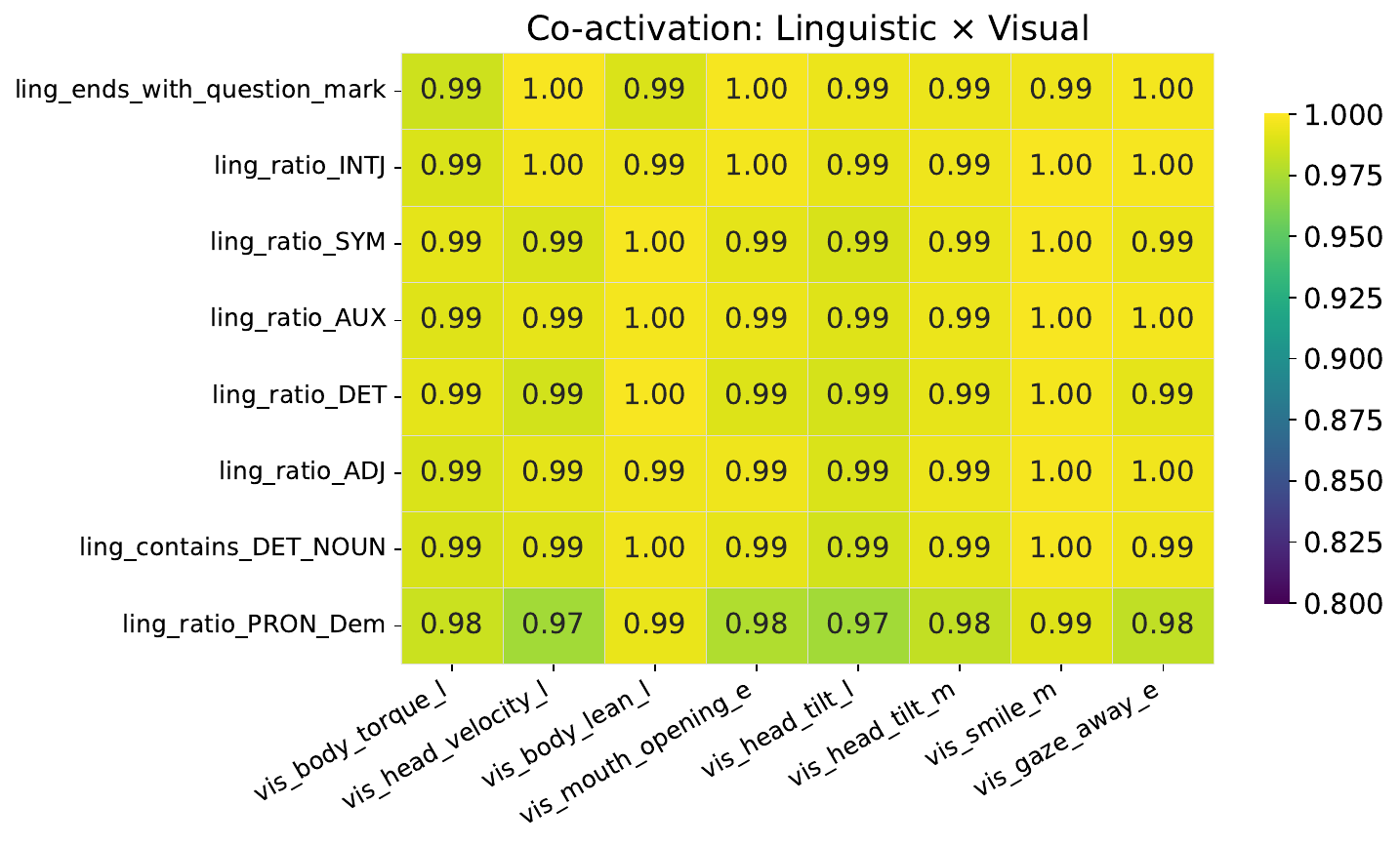}
\caption{CABB-S's Linguistic vs Visual co-activation heatmap}
\label{fig:cabbs_lin_vis_syn}
\end{center}
\end{figure}

\begin{figure}[h]
\begin{center}
\includegraphics[width=\columnwidth]{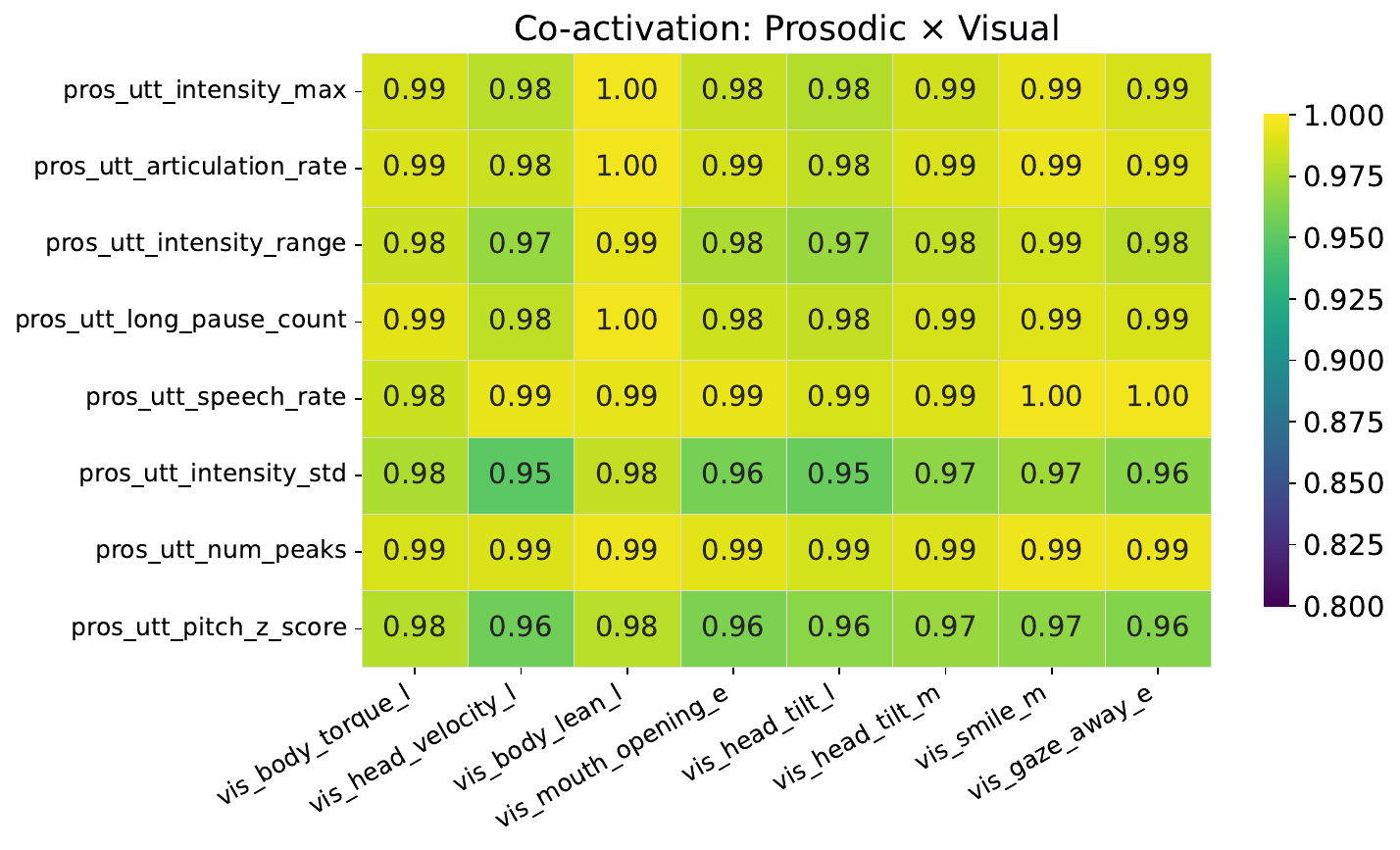}
\caption{CABB-S's Prosodic vs Visual co-activation heatmap}
\label{fig:cabbs_pros_vis_syn}
\end{center}
\end{figure}

For CABB-S, both linguistic-visual (Figure~\ref{fig:cabbs_lin_vis_syn}) and prosodic-visual (Figure~\ref{fig:cabbs_pros_vis_syn}) co-activation scores remain consistently high (0.97—1.00 and 0.95-1.00, respectively), with no independently activated visual feature comparable to \textbf{vis\_hold\_release}. Overall, these results suggest that the model consistently relies on coordinated linguistic, prosodic, and visual cues when recognizing repair initiation in CABB-S.

\section{Conclusion \& Future Works}

This work presented the first computational study to investigate visual behaviors for automatic  OIR repair initiation detection, introducing a Conversation Analysis (CA)-grounded visual feature representation for interpretable multimodal modeling. We evaluated the proposed approach on two corpora with different languages and  interactional contexts: NOXI (French, screen-mediated) and CABB-S (Dutch, face-to-face, task-oriented).

Results demonstrate that incorporating visual modality consistently improves OIR detection and classification over text-audio baselines and substantially outperforms a zero-shot multimodal large language model. Furthermore, cross-linguistic and cross-context evaluation reveals that visual behaviors contribute differently across interaction settings. Visual cues primarily improve detection in CABB-S, with richer body-level information, whereas they contribute more to repair-type classification in NOXI, where facial behaviors help disambiguate lexically similar repair expressions. Feature-level analyses further provide insights into model behavior, such as facial displays and interactional cues in the beginning of the segment are most informative in NOXI, while body posture and middle- to late-phase behaviors contribute more in CABB-S. Co-activation analysis further reveals how visual behaviors interact with linguistic and prosodic signals across different conversational settings. These findings demonstrate that CA-grounded visual features provide both performance gains and interpretable characterizations of repair behavior, helping bridge interactional theory and computational modeling. Future work will evaluate the proposed framework on larger conversational corpora and investigate its deployment for real-time repair recognition in conversational agents.

\begin{acks}
We thank the anonymous reviewers for their constructive feedback. Data were provided (in part) by the Radboud University, Nijmegen, The Netherlands. This work has been supported by the Paris Île-de-France Région in the framework of DIM AI4IDF. It was also partially funded by the ANR-23-CE23-0033-01 SINNet project.
\end{acks}

\section*{Safe and Responsible Innovation Statement}

This work uses existing annotated corpora collected under ethical oversight, without gathering new human data. Automatic OIR detection can enhance dialogue systems and assistive technologies by facilitating recovery from communication breakdowns, particularly for users with hearing or language difficulties. However, visual features (e.g., gaze, facial expressions, body movement) raise greater privacy concerns than text or audio, particularly if used without user consent. While we evaluate across French and Dutch corpora, performance differences across contexts highlight potential risks of unequal reliability across languages and interaction styles. We therefore emphasize the need for transparency and validation when visual data is used.

\bibliographystyle{ACM-Reference-Format}
\bibliography{sample-base}

\appendix









\end{document}